# Sparse Support Vector Infinite Push

**Alain Rakotomamonjy**                                            ALAIN.RAKOTO@INSA-ROUEN.FR

LITIS EA 4108, Université de Rouen Avenue de l'université, 76800 Saint Etienne du Rouvray, France

## Abstract

In this paper, we address the problem of embedded feature selection for ranking on top of the list problems. We pose this problem as a regularized empirical risk minimization with $p$-norm push loss function ($p = \infty$) and sparsity inducing regularizers. We leverage the issues related to this challenging optimization problem by considering an alternating direction method of multipliers algorithm which is built upon proximal operators of the loss function and the regularizer. Our main technical contribution is thus to provide a numerical scheme for computing the infinite push loss function proximal operator. Experimental results on toy, DNA microarray and BCI problems show how our novel algorithm compares favorably to competitors for ranking on top while using fewer variables in the scoring function.

## 1. Introduction

Learning to rank is a supervised learning problem which objective is to estimate a scoring function from training examples. That function is expected to define a partial order on the examples by scoring relevant instances higher than the non-relevant ones. Examples of applications in which ranking is central are information retrieval (Chapelle & Keerthi, 2010), drug discovery (Agarwal et al., 2010).

Many machine learning algorithms have been proposed for learning ranking functions. Some of them aim at optimizing a pairwise ranking criterion using exponential loss, like the RankBoost of Freund et al. (2003) or using Hinge loss (Joachims, 2002). Methods based on decision trees have also been investigated (Clémencon & Vayatis, 2009). While these methods have shown their interests, they may not be optimally targeted at a specific goal which is getting an accurate ranking at the top of the list. Indeed, for instance, in information retrieval, one usually wants to have the most relevant documents as possible at the top of the list, while getting an accurate pairwise ordering in other part of the list is not of great importance. For this reason, several recently proposed algorithms focus on correctly ranking the best instances (Rudin, 2009; Agarwal, 2011).

For most of the works we described above, the scoring function is a linear function of the form $f(\mathbf{x}) = \mathbf{w}^\top \mathbf{x}$, where $\mathbf{w}$ is the weight vector that has to be learned. The resulting weight vector $\mathbf{w}$ is usually a non-sparse vector, which means that all features will be considered in the scoring function even though they are non-informative. Hence, similarly to other supervised learning paradigms, ranking methods may also benefit from feature selection as keeping only few features in the scoring function may improve performances as well as reducing prediction time. To the best of our knowledge, very few works have addressed the problem of feature selection in ranking (Geng et al., 2007) and in ranking on top of the list problems. Naturally, algorithms like the SVM-RFE or methods based on sparsity-inducing norms like the $\ell_1$ norm can be easily extended to ranking algorithms. But developing algorithms for embedded feature selection becomes more challenging when loss functions related to ranking on top have to be considered. This is the challenge we want to address in this paper and as far as we know, this is the first paper proposing embedded feature selection using sparsity inducing norms for ranking on top of the list.

We focus on the recent $p$-norm push loss function introduced by Rudin (2009) and more specifically, on the case where $p = \infty$, denoted as an infinite push framework by Agarwal (2011). In this latter work, Agarwal (2011) has also proposed a support vector like algorithm, named as support vector Infinite Push, which has been proved to perform better than competitors when the goal is to maximize the number of relevant instances on top of the list. Here, we provide a novel algorithm for solving the regularized empirical risk minimization related to the support vector Infinite Push loss function that can also handle sparsity inducing regularizers. While typical methods dealing with sparsity suppose that the loss function is smooth (Bach et al., 2011), our optimization problem is challenging since both the loss function and the regularizer can be non-differentiable. We propose to overcome this issue by proposing an alternating direction method of multipliers (ADMM), which is wrapped around the computation of the infinite push loss function proximal operator in addition to the one of the regularizer.

Globally, the paper provides contributions to the state of the art in several points: (a) it proposes a numerical scheme for computing the proximal operator of the support vector infinite push loss function, (b) it shows that the optimization framework we consider offers some theoretical guarantees on the uniqueness of the problem minimizer, property that is not always insured by Agarwal's algorithm, (c) it is the first paper showing that $p$-norm push ranking algorithms can also embed feature selection through the use of sparsity inducing norms. It demonstrates that ranking on top applications can also benefit from feature selection either by improved performances or reduced prediction time.

The paper is organized as follows. Section 2 introduces the global framework for ranking on top of the list as well as the optimization related to sparse support vector infinite push. In Section 3, the ADMM-based algorithm proposed for solving the problem and the numerical scheme for computing the proximal operator of the infinite push loss are presented. Experimental results are described in Section 4 while conclusion is in Section 5.

## 2. Infinite Push framework

In this section, we introduce the Infinite Push loss function and the support vector Infinite Push optimization problem we are interested in. Existence and uniqueness of solutions to the problem are also discussed.

### 2.1. Infinite Push loss function

We limited ourselves to the case of bipartite ranking problem which goal is to learn a function that, given a training set $\{\mathbf{x}_i\}_{i=1}^{\ell}$, $\mathbf{x}_i \in \mathbb{R}^d$, with $m$ positive and $n$ negative examples, gives higher scores to positive examples than to negatives ones. Learning such a function can be cast into an empirical regularized risk minimization framework, where the loss function related to the risk is designed so as to favor higher scores for positive examples. Typically, in such a context, the loss function focuses on the average pairwise scoring losses and it can be written as :

$$L(f(\cdot), S) = \frac{1}{mn} \sum_{i,j} I_{f(\mathbf{x}_i^+) \leq f(\mathbf{x}_j^-)}$$

where $I.$ is the indicator function, $S$ is a set of examples with known labels and $f(\cdot)$ is the scoring function that we want to evaluate. Several extensions of this loss function have been recently considered in order to provide more importance to errors made on top of the lists, for instance by weighting the pairwise loss (Usunier et al., 2009) or by replacing the mean with some more appropriate functions. For this purpose, Rudin (2009) has introduced the Infinite Push loss function

$$L(f(\cdot), S)_\infty = \max_j \sum_i I_{f(\mathbf{x}_i^+) \leq f(\mathbf{x}_j^-)} \qquad (1)$$

which gets smaller as the negative example with highest score is assigned a small score. This loss function is the one on which we have focused our interest.

### 2.2. Support Vector Infinite Push

We can now define the empirical risk minimization (ERM) framework used for learning the scoring function $f(\cdot)$ that we have chosen to be linear so that $f(\mathbf{x}) = \mathbf{w}^\top \mathbf{x}$. the loss function given in Equation (1) is non-convex and different convexifications proposed in the literature have led to different ERM frameworks and algorithms. We can mention for instance the relaxation by means of exponential loss that yield to boosting-like algorithm (Rudin, 2009). If Hinge loss is used as a convex relaxation then we get the following Support Vector like optimization problem :

$$\min_{\mathbf{w}} \quad \lambda \Omega(\mathbf{w}) + \max_{1 \leq j \leq n} \left( \frac{1}{m} \sum_{i=1}^m \left(1 - \mathbf{w}^\top (\mathbf{x}_i^+ - \mathbf{x}_j^-)\right)_+ \right) \qquad (2)$$

where $\Omega(\mathbf{w})$ is some regularization term and the function $(u)_+ = u$ if $u > 0$ and $0$ otherwise.

For $\Omega(\mathbf{w}) = \frac{\lambda}{2}\|\mathbf{w}\|^2$, Agarwal has proposed an algorithm for solving the dual of this problem, which is :

$$\begin{aligned} \min_{\alpha_{i,j}} \quad & \frac{1}{2} \sum_{i,j} \sum_{k,l} \alpha_{i,j} \alpha_{k,l} (\mathbf{x}_i^+ - \mathbf{x}_j^-)^\top (\mathbf{x}_k^+ - \mathbf{x}_l^-) - \sum_{i,j} \alpha_{i,j} \\ \text{st} \quad & \sum_j \max_i(\alpha_{i,j}) \leq \frac{1}{\lambda m} \\ & \alpha_{i,j} \geq 0 \end{aligned}$$
$$(3)$$

which has smooth quadratic objective function under some mixed-norm constraints over the dual variables $\alpha_{i,j}$. The algorithm is based on a nice and clever gradient projection algorithm (Agarwal, 2011).

In this paper, we focus our effort on this support vector infinite push problem and propose a novel optimization algorithm for solving it when $\Omega(\mathbf{w})$ is some non-differentiable sparsity-inducing regularizer so as to perform feature selection in a top-ranking learning problem. For this purpose, we investigate an algorithm that directly solves the primal problem in Equation (2).

However, before delving into the details of the algorithm, we discuss, in what follows the existence and uniqueness of the solution of the primal problem (2) based on some classical results on convex analysis:

**Proposition 1.** *(a) For any convex regularization term $\Omega(\mathbf{w})$ that is lower semi-continuous and and coercive, problem (2) admits at least one solution. (b) For any strictly convex, lower semi continuous and coercive regularization term, problem (2) admits an unique solution.*

We omit the proof of these two propositions since they are rather direct consequences of some well known results on the minimization of composite non-smooth functions (Combettes & Pesquet, 2007). Instead, we prefer to bring to light some properties of the primal problem compared to the dual one that are consequences of these propositions :

An interesting point is that for $\Omega(\mathbf{w}) = \frac{\lambda}{2}\|\mathbf{w}\|^2$, point 2 of the proposition guarantees uniqueness of solution since $\Omega$ satisfies all required properties. Conversely, when considering the dual problem (3) as in Agarwal (2011), this property may be lost. Indeed, it can be easily shown that when the dimensionality of the problem $d$ is smaller than $m \cdot n$, the Hessian of the dual objective function is only positive semi-definite. We remark that even for small-scale high-dimensional learning problem, the condition $d < m \cdot n$ can be rapidly reached making optimization in the primal theoretically more sound.

Uniqueness of the solution for $\ell_1$ norms or mixed-norms are more involved and we have left these analyses for future works.

## 3. Algorithm for sparse Support Vector Infinite Push

In this section, we show how we leverage the issues raised by the non-smooth objective function in problem (2) and we describe in details the ADMM algorithm we propose for solving the sparse support vector infinite push problem.

### 3.1. Deriving ADMM formulation

Before delving into the derivations, we want to mention that Douglas-Rachford splitting algorithm is tailored for minimizing the sum of two non-smooth objective functions. However, the presence of the design matrix will add some linear constraint on the problem, making it easier to address through an ADMM framework. For this purpose, we rewrite the optimization problem (2) as the following linearly-constrained problem :

$$\min_{\mathbf{w},\mathbf{a}} \quad \Omega(\mathbf{w}) + \max_{1 \leq j \leq n} \left( \frac{1}{m} \sum_i (a_{i,j})_+ \right) \quad (4)$$
$$a_{i,j} = 1 - \mathbf{w}^T(\mathbf{x}_i^+ - \mathbf{x}_j^-)$$

where $\Omega(\mathbf{w})$ can be any sparsity inducing norm like the $\ell_1$ norm, any mixed-norm (Bach et al., 2011) or the classical $\ell_2$ regularization term. Then, by properly defining the matrix $\mathbf{X}$ (which rows are of the form $(\mathbf{x}_i^+ - \mathbf{x}_j^-)^T$), the vector $\mathbf{a}$ and the function $g(\mathbf{a}) = \max_j \left(\frac{1}{m}\sum_i \max(a_{i,j}, 0)\right)$ we yield the following reformulation :

$$\begin{aligned} \min_{\mathbf{w},\mathbf{a}} \quad & \Omega(\mathbf{w}) + g(\mathbf{a}) \\ & \mathbf{X}\mathbf{w} + \mathbf{a} - \mathbf{1} = 0 \end{aligned} \quad (5)$$

The augmented Lagrangian related to this problem is

$$\begin{aligned} \mathcal{L}(\mathbf{w}, \mathbf{a}, \delta, \mu) &= \Omega(\mathbf{w}) + g(\mathbf{a}) + \delta^\top(\mathbf{X}\mathbf{w} + \mathbf{a} - \mathbf{1}) \\ &\quad + \frac{\mu}{2}\|\mathbf{X}\mathbf{w} + \mathbf{a} - \mathbf{1}\|^2 \end{aligned}$$

where $\delta$ is a vector of Lagrangian multipliers related to the equality constraint and $\mu$ is a parameter weighting the quadratic penalty. After rearranging the terms, one can show that the augmented Lagrangian is

$$\mathcal{L}(\mathbf{w}, \mathbf{a}, \gamma) = \Omega(\mathbf{w}) + g(\mathbf{a}) + \frac{\mu}{2}\|\mathbf{X}\mathbf{w} + \mathbf{a} - \mathbf{1} + \gamma\|^2$$

where $\gamma = \frac{\delta}{\mu}$. The alternating direction method of multipliers that solves our original problem (4) looks for a saddle point of the augmented Lagrangian by solving alternatively at iteration $k$ the following problems :

$$\begin{aligned} \mathbf{w}^{k+1} &= \arg\min_{\mathbf{w}} \mathcal{L}(\mathbf{w}, \mathbf{a}^k, \gamma^k) & (6) \\ \mathbf{a}^{k+1} &= \arg\min_{\mathbf{a}} \mathcal{L}(\mathbf{w}^{k+1}, \mathbf{a}, \gamma^k) & (7) \\ \gamma^{k+1} &= \gamma^k + \mathbf{X}\mathbf{w}^{k+1} + \mathbf{a}^{k+1} - \mathbf{1} & (8) \end{aligned}$$

All the challenges of the algorithm now resides essentially in the resolution of these problems.

### 3.2. Solving problem (6)

The optimization problem related to $\mathbf{w}$ can be restated as

$$\min_{\mathbf{w}} \frac{1}{2}\|\mathbf{X}\mathbf{w} - \mathbf{s}\|_2^2 + \frac{1}{\mu}\Omega(\mathbf{w})$$

with $\mathbf{s}$ being $1 - \mathbf{a}^k - \gamma^k$. Depending on the form of $\Omega(\mathbf{w})$, this problem becomes a ridge regression problem for $\Omega(\mathbf{w}) = \frac{\lambda}{2}\|\mathbf{w}\|_2$, a Lasso when $\Omega(\mathbf{w}) = \lambda\|\mathbf{w}\|_1$, or another (probably known) problem if a different regularization term is considered.

For sparsity-inducing regularizers (e.g $\ell_1$ norm), the problem has to be solved numerically and thus, each iteration of the ADMM approach involves the resolution of a Lasso. Depending on the Lasso algorithm used, one can highly benefit from warm-starting the solution since between two consecutive ADMM iteration, the second member $\mathbf{s}$ is not expected to vary a lot.

For the $\ell_2$ norm regularizer, the solution has a closed-form solution

$$\mathbf{w}^{k+1} = (\mathbf{X}^\top \mathbf{X} + \frac{\lambda}{\mu}\mathbf{I})^{-1}(\mathbf{X}^\top \mathbf{s})$$

In some situations, when the dimensionality of the problem is large, it may be more efficient to numerically solve this linear system by means of a conjugate gradient descent approach.

### 3.3. Solving problem (7)

Now supposing that $\mathbf{w}$ and the Lagrangian multipliers $\gamma$ are fixed in the Lagrangian, the optimization problem related to (7) boils down to be :

$$\mathbf{a}^{k+1} = \arg\min_{\mathbf{a}} g(\mathbf{a}) + \frac{\mu}{2}\|\mathbf{a} - \mathbf{s}\|_2^2 \qquad (9)$$

with $\mathbf{s}$ being $1 - \gamma^k - \mathbf{X}\mathbf{w}^{k+1}$. We note that by definition, $\mathbf{a}^{k+1}$ is the result of $\frac{1}{\mu}g(\cdot)$ proximal operator applied to the vector $\mathbf{s}$ (Combettes & Pesquet, 2010). Now, let us look into more details at this problem. The most challenging part of it comes from the two nested max functions defining $g(\cdot)$. In order to overcome part of the issues, we propose to use the doubling trick and rewrites the minimization problem as :

$$\begin{array}{ll} \min_{\mathbf{a}^+,\mathbf{a}^-} & \frac{1}{2}\|\mathbf{a}^+ - \mathbf{a}^- - \mathbf{s}\|_2^2 + \max_j\left(\frac{1}{m\mu}\sum_i \mathbf{a}^+\right) \\ \text{st} & \mathbf{a}^+ \geq 0, \mathbf{a}^- \geq 0 \end{array}$$
(10)

with $\mathbf{a} = \mathbf{a}^+ - \mathbf{a}^-$. Now, since the regularization term and the constraints are decoupled in $\mathbf{a}^+$ and $\mathbf{a}^-$, we suggest to solve problem (10) by means a of block-coordinate descent (BCD) algorithm that starts from some positive random vectors and alternatively optimize over $\mathbf{a}^+$ then $\mathbf{a}^-$ keeping the other vector fixed. Before providing algorithmic details, we state here a proposition based on the work of Tseng (2001) that guarantees the soundness of the BCD algorithm.

**Proposition 2.** *Let us define* $f_0(\mathbf{a}^+,\mathbf{a}^-) = \frac{1}{2}\|\mathbf{a}^+ - \mathbf{a}^- - \mathbf{s}\|_2^2$, $f_1(\mathbf{a}^+) = \max_j\left(\frac{1}{m\mu}\sum_i \mathbf{a}^+\right) + I_{\mathbf{a}^+\geq 0}$, $f_2(\mathbf{a}^-) = I_{\mathbf{a}^-\geq 0}$, *the sequence generates by the BCD method by alternatively optimizing over* $\mathbf{a}^+$ *and* $\mathbf{a}^-$ *converges towards the minimum of Problem 10.*

*Proof.* It is easy to see that the objective function $f(\mathbf{a}^+,\mathbf{a}^-)$ of Problem 10 is $f(\mathbf{a}^+,\mathbf{a}^-) = f_0(\mathbf{a}^+,\mathbf{a}^-) + f_1(\mathbf{a}^+,\mathbf{a}^-) + f_2(\mathbf{a}^+,\mathbf{a}^-)$. Besides, $f_0$ is continuous on its domain and coercive, $f(\cdot,\cdot)$ is convex with respect to any of its parameter with the other fixed and the functions $f_i(\cdot), i = \{0,1,2\}$ are lower semi-continuous. Owing to all these properties, applying Theorem 5.1 of Tseng (2001) concludes the proof. $\square$

Now, we are interested in solving each coordinate descent of Equation (10). When considering minimizing over $\mathbf{a}^-$ with $\mathbf{a}^+$ fixed, the problem is rather simple since it boils down to be a projection of $-\mathbf{s} + \mathbf{a}^+$ on the positive quadrant. Hence, we have the following closed-form solution for each component of $\mathbf{a}^-$ :

$$a_k^- = \begin{cases} -s_k + a_k^+ & \text{if } -s_k + a_k^+ \geq 0 \\ 0 & \text{otherwise} \end{cases}$$

We can now focus on the other alternate problem

$$\begin{array}{ll} \min_{\mathbf{a}^+} & \frac{1}{2}\|\mathbf{a}^+ - \mathbf{b}\|_2^2 + \max_j\left(\frac{1}{m\mu}\sum_{i \in G_i} a_i^+\right) \\ \text{st} & \mathbf{a}^+ \geq 0 \end{array}$$

with $\mathbf{b} = \mathbf{a}^- + \mathbf{s}$ and $G_i$ being the indices of elements of $\mathbf{a}$ coupling the negative example $\mathbf{x}_j$ with positive examples. Interestingly, owing to the positiveness of $\mathbf{a}^+$ and by replacing the constraint in the objective value this problem is equivalent to

$$\min_{\mathbf{a}^+} \quad \frac{1}{2}\|\mathbf{a}^+ - \mathbf{b}\|_2^2 + \max_j\left(\frac{1}{m\mu}\sum_{i \in G_i} |a_i^+|\right) + I_{\mathbf{a}^+\geq 0}$$
(11)

We can note here that the solution of this problem occurs at

$$\mathbf{a}^{+\star} = \text{prox}_{I_{\cdot \geq 0}+\Omega^\dagger}(\mathbf{b})$$

with $I_{\cdot \geq 0}$ being the indicator on the positive quadrant and $\Omega^\dagger(\mathbf{u}) = \max_j\left(\frac{1}{m\mu}\sum_{i \in G_i} |u_i^+|\right)$ which is a mixed $\ell_\infty - \ell_1$ norm on $\mathbf{u}$. The proximal operator $\text{prox}_{I_{\cdot \geq 0}+\Omega}(\mathbf{b})$ is non-trivial and needs to be computed numerically. For this purpose, we have applied a Douglas-Rachford algorithm which can handle the minimization of the sum of two non-smooth convex functions $f_1$ and $f_2$. The following proposition makes this explicit :

**Proposition 3.** *(Combettes & Pesquet, 2010) Let $f_1$ and $f_2$ be two convex lower semi-continuous functions of $\mathbb{R}^d$ such that the intersection of their domain relative interiors is not empty and such that $f_1(\cdot) + f_2(\cdot)$ is coercive. Set $\mathbf{v}_0 \in \mathbb{R}^d$ and build $\mathbf{u}_n$ for $n \geq 0$ as*

$$\begin{aligned} \mathbf{u}_n &= prox_{\gamma f_2}(\mathbf{v}_n) \\ \mathbf{v}_{n+1} &= \mathbf{v}_n + \eta\left(prox_{\gamma f_1}(2\mathbf{u}_n - \mathbf{v}_n) - \mathbf{u}_n\right) \end{aligned} \quad (12)$$

*with $\gamma > 0$ and $\eta \in ]0, 2[$, then every sequence $\{\mathbf{u}_n\}$ generated by this algorithm converges towards a minimizer of $f_1 + f_2$.*

Hence, a direct application of this algorithm to our problem given in Equation (11) with $f_2(\mathbf{v}) = \frac{1}{2}\|\mathbf{v} - \mathbf{b}\|_2^2 + I_{\mathbf{v} \geq 0}$ and $f_1(\mathbf{v}) = \max_j \left(\frac{1}{m\mu} \sum_{i \in G_i} |v_i|\right)$ leads to the minimizer of Equation (11). Now the remaining question is : what are the proximal operators of $\gamma f_1$ and $\gamma f_2$?

For $\gamma f_2$, we have to solve the problem

$$\text{prox}_{\gamma f_2}(\mathbf{v}_n) = \arg\min_{\mathbf{z}} \frac{1}{2}\|\mathbf{z} - \mathbf{v}_n\|_2^2 + \frac{\gamma}{2}\|\mathbf{z} - \mathbf{b}\|_2^2 + I_{\mathbf{z} \geq 0}$$

which solution can be easily proven to be

$$\text{prox}_{\gamma f_2}(\mathbf{v}_n) = P_C\left(\frac{\mathbf{v}_n + \gamma \mathbf{b}}{1 + \gamma}\right)$$

with $P_C$ being the projection on the positive quadrant. Now, regarding $\gamma f_1$, we look for

$$\text{prox}_{\gamma f_1}(\mathbf{v}) = \arg\min_{\mathbf{z}} \frac{1}{2}\|\mathbf{z} - \mathbf{v}\|_2^2 + \frac{\gamma}{m\mu} \max_j \left(\sum_{i \in G_i} |z_i|\right) \quad (13)$$

which is the proximal operator of a $\ell_\infty - \ell_1$ mixed norm. For solving this problem, we use classical result from convex analysis and proximal operator (Combettes & Wajs, 2005; Sra, 2011), which states that the proximal operator of a norm $\|\cdot\|$ is

$$\text{prox}_{\tau\|\cdot\|}(\mathbf{u}) = \mathbf{u} - \Pi_{\|\cdot\|_* \leq \tau}(\mathbf{u})$$

with $\Pi_{\|\cdot\|_* \leq \tau}(\mathbf{u})$ being the projection of $\mathbf{u}$ on the $\tau$-radius ball of the dual norm $\|\cdot\|_*$. Then, since the dual norm of $\ell_\infty - \ell_1$ norm is the $\ell_1 - \ell_\infty$ norm, we have

$$\text{prox}_{\gamma f_1}(\mathbf{v}) = \mathbf{v} - \Pi_{\|\cdot\|_{1,\infty} \leq \frac{\gamma}{m\mu}}(\mathbf{v}) \quad (14)$$

This problem is easily tractable since projection of vector on a $\ell_1 - \infty$ ball has been recently studied and several efficient algorithms proposed (Quattoni et al., 2009; Sra, 2011).

**Algorithm 1** ADMM approach for primal infinite push.
1: Input : $\mathbf{X}$ : matrix of pairwise difference of examples, $\lambda$ regularization term
2: set $\mu > 0$, $k = 0$
3: initialize $\mathbf{a}^k$ and $\gamma^k$ to vectors of 0.
4: **repeat**
5: $\quad \mathbf{s} = \mathbf{1} - \mathbf{a}^k - \gamma^k$
6: $\quad \mathbf{w}^{k+1} = \arg\min_{\mathbf{w}} \frac{1}{2}\|\mathbf{X}\mathbf{w} - \mathbf{s}\|_2^2 + \lambda \Omega(\mathbf{w})$
7: $\quad \mathbf{s} = \mathbf{1} - \gamma^k - \mathbf{X}\mathbf{w}^{k+1}$
8: $\quad$ set $\mathbf{a}^+ = 0$ and $\mathbf{a}^- = 0$
9: $\quad$ **repeat**
10: $\quad\quad \mathbf{a}^- = \max(-\mathbf{s} + \mathbf{a}^+, 0)$
11: $\quad\quad$ set $\mathbf{b} = \mathbf{a}^- + \mathbf{s}$ and $\mathbf{v}_0 = 0$
12: $\quad\quad$ **repeat**
13: $\quad\quad\quad \mathbf{u}_n = \frac{1}{1+\gamma} \max(\mathbf{v}_n + \gamma \mathbf{b}, 0)$
14: $\quad\quad\quad \mathbf{v}_{n+1} = \mathbf{v}_n + \eta(\mathbf{u}_n - \mathbf{v}_n - \Pi_{\|\cdot\|_{1,\infty}}(2\mathbf{u}_n - \mathbf{v}_n))$
15: $\quad\quad$ **until** convergence is met
16: $\quad\quad \mathbf{a}^+ = \mathbf{u}_n$
17: $\quad$ **until** convergence is met
18: $\quad \mathbf{a}^{k+1} = \mathbf{a}^+ - \mathbf{a}^-$
19: $\quad \gamma^{k+1} = \gamma^t + \mathbf{X}\mathbf{w}^{k+1} + \mathbf{a}^{k+1} - \mathbf{1}$
20: $\quad k \leftarrow k + 1$
21: **until** condition

### 3.4. Convergence analysis

Convergence of Algorithm 1, for solving the primal infinite push problem builds upon classical convergence results of ADMM or Douglas-Rachford splitting algorithm (Eckstein & Bertsekas, 1992). Indeed, a direct application of Theorem 8 in that paper tells us that our algorithm converges for any $\mu > 0$, as long as the matrix $\mathbf{X}$ has full column rank (condition that is satisfied by most non-degenerate problems for which $d < m \cdot n$) and that the computation errors of problem (6) and problem (7) are summable. Practically, this latter condition means that the convergence criterion on these two problems should become tighter and tighter as the iterations go. However, in our implementation, these stopping criteria have been kept fixed but still no empirical problem of convergence has been noticed.

### 3.5. Computational complexity

The two most computationally demanding part of our algorithm for sparse infinite push is the Lasso problem that has to be solved at each iteration and the projection on the $\ell_1 - \ell_\infty$ ball. For the Lasso, there exists efficient algorithms that scale linearly with the number of training examples. We can, for instance, mention the SpaRSa algorithm of Figueiredo et al. (Figueiredo et al., 2007). Similarly, the projection on the $\ell_1 - \ell_\infty$ ball of Quattoni et al. (2009) has a complexity of $\mathcal{O}(n \log n)$ in the size of the vector to project. Hence, in our case, since the number of examples in the Lasso and the size of the vector to project are both $m \cdot n$, we end up with an algorithm which complexity is $\mathcal{O}(m \cdot n \log m \cdot n)$.

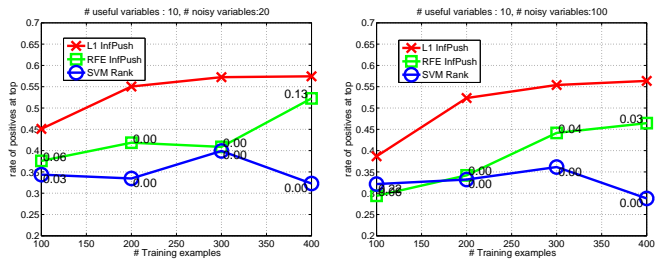

Figure 1. Rate of positives at top of the list of when the number of discriminative features is 10 and the number of noisy variables is : left) 20. right) 100.

As a comparison, a plain implementation of RankSVM would also lead to a complexity that is linear with respects to the number of pairwise examples (Chapelle & Keerthi, 2010). However, since RankSVM is easier to deal with as the loss function can be made differentiable we expect RankSVM to have a better constant.

## 4. Experiments

Our objective here is to provide empirical evidences that our method can be beneficial in problems with noisy or redundant features compared to an infinite push approach that considers all the features. We also show that when compared to other feature selection methods like recursive feature elimination (RFE) in an infinite push context, our embedded approach based on sparsity-inducing norm provides better accuracy on top of the list.

Note that we have not compared our methods to other ranking algorithms, except SVMRank (Chapelle & Keerthi, 2010), since Agarwal (2011) has already shown the superiority of the infinite push model on other methods for ranking positive instances on top of the list and because these methods such as SVMMAP does not have their sparse counterpart in the literature.

### 4.1. Toy problem

On this problem, we compare the efficiency of using an $\ell_1$ sparsity-inducing norm to recursive-feature elimination for reducing the influence of noisy variables in an infinite push framework. Our RFE implementation follows exactly the same procedure as the one used for SVM RFE (Guyon et al., 2002), but replacing the SVM with the infinite push algorithm as proposed by Agarwal (2011). This infinite push RFE bears strong resemblance with the backward elimination of Geng et al. (2007). For a baseline comparison, we have also included an $\ell_1$ SVM Rank.

The toy problem is a binary classification problem in $\mathbb{R}^d$ with evenly distributed classes. Among these $d$ variables, only $r$ of them define a subspace of $\mathbb{R}^d$ in which classes can be discriminated. For these $r$ relevant variables, the two classes follow a Gaussian pdf with mean respectively $\mu$ and $-\mu$ and covariance matrices randomly drawn from a Wishart distribution. $\mu$ has been randomly drawn from $\{-1, +1\}^r$. The other $d - r$ non-relevant variables follow an $i.i.d$ Gaussian probability distribution with zero mean and unit variance for both classes. We have respectively sampled $n$ and $n_t$ number of examples for training and testing. For some experiments, $n$ is varying, but we have always set $n_t = 1000$. Before learning, the training set has been normalized to zero mean and unit variance and the test sets have been rescaled accordingly. Hyperparameters of all methods have been chosen as those maximizing performance on a validation set obtained by random $70\% - 30\%$ split of the training set examples.

Averaged results over 20 trials are depicted on Figure 1 which plots the rate of positive on top of the list defined as $\frac{\text{\#pos. on top}}{m}$, with respects to varying number of training examples for fixed number of features. We note that our $\ell_1$ support vector infinite push significantly outperforms other competitors, in most cases with a p-value of a Wilcoxon signed rank test lower than 0.05 (the numbers besides the markers). We can also remark that unlike the infinite push approach, SVM Rank does not necessarily improve its performances on the top as the number of examples increases. This is unsurprising as SVM Rank aims at optimizing average ranking.

Figure 2 depicts the precision and the F-measure of the different algorithms for retrieving the true variables. We remark that the RFE infinite push performs very good with respects to the F-measure. However, the use of the $\ell_1$ norm yields to a better precision : more relevant variables are selected at the expense of selecting some irrelevant ones. This is a well known issue of the $\ell_1$ norm that can be overcome using an adaptive approach (Zou, 2006).

An empirical illustration of the computational complexity of our algorithm as well as the one of a sparse

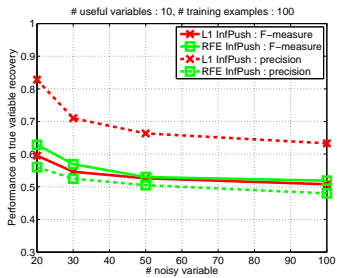

Figure 2. Evolution of the F-measure on retrieved features as a function of the number of noisy ones for 100 training examples.

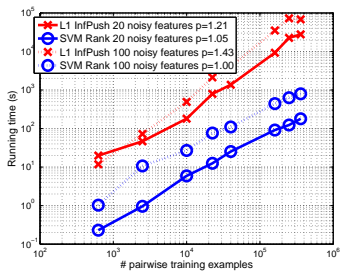

Figure 3. Examples of running time as the number of pairwise training examples increases for a $\ell_1$ Infinite Push and an $\ell_1$ RankSVM with (solid) 20 and (dotted) 100 noisy feature. $p$ is empirical exponent of the computational complexity.

SVM Rank is reported in Figure 3. We can highlight that both algorithms have an empirical exponent complexity of about 1 with respects to the number of pairwise training examples.

### 4.2. Real-world problems

We also also carried out experiments on some real-world datasets. These datasets are essentially related to DNA microarray analysis (colon,yeast), or comes from the UCI dataset repository (ionosphere, sonar, spectf, wpbc), as well as P300 based BCI speller. The same pre-processing as for the toy dataset has been applied to these real ones.

We have compared a plain infinite push method that does not perform embedded variable selection and a $\ell_1$ SVM Rank to our sparse $\ell_1$ infinite push. Comparison criteria are the rate of positive examples ranked on top and the number of variables used by the scoring functions. Averaged results over 10 iterations have been reported in Table 1. Results clearly shows that our $\ell_1$ infinite push model is the model that achieves the best compromise between accuracy on top of the list and variable selection. Indeed, for all datasets, performances on top are statistically equivalent whereas sparse infinite push uses significantly fewer variables in most of the cases. A reduction of a factor 20 or 8 can respectively be achieved with respects to the original number of variables or the number of variables selected by SVM Rank.

| Data | Algo | top | # var |
|---|---|---|---|
| colon | $\ell_1$ IP | $0.41 \pm 0.3$ | $\mathbf{68.20 \pm 25.4}$ |
| $d = 2000$ | $\ell_2$ IP | $0.36 \pm 0.2$ | - |
| (43-19) | $\ell_1$ Rank | $0.40 \pm 0.3$ | $572.40 \pm 320.1$ |
| yeast | $\ell_1$ IP | $0.85 \pm 0.2$ | $\mathbf{25.90 \pm 5.3}$ |
| $d = 79$ | $\ell_2$ IP | $\underline{0.53 \pm 0.4}$ | - |
| (124-208) | $\ell_1$ Rank | $0.86 \pm 0.2$ | $64.20 \pm 2.8$ |
| sonar | $\ell_1$ IP | $0.44 \pm 0.2$ | $\mathbf{23.70 \pm 6.5}$ |
| $d = 60$ | $\ell_2$ IP | $0.48 \pm 0.3$ | |
| (187-208) | $\ell_1$ Rank | $0.39 \pm 0.2$ | $59.80 \pm 0.4$ |
| wpbc | $\ell_1$ IP | $0.18 \pm 0.3$ | $26.00 \pm 12.8$ |
| $d = 33$ | $\ell_2$ IP | $0.34 \pm 0.3$ | - |
| (174-194) | $\ell_1$ Rank | $0.34 \pm 0.1$ | $32.70 \pm 0.7$ |
| iono | $\ell_1$ IP | $0.64 \pm 0.2$ | $\mathbf{15.00 \pm 5.1}$ |
| $d = 33$ | $\ell_2$ IP | $0.66 \pm 0.1$ | - |
| (245-351) | $\ell_1$ Rank | $0.69 \pm 0.1$ | $33.00 \pm 0.0$ |
| spectf | $\ell_1$ IP | $0.14 \pm 0.1$ | $40.70 \pm 4.4$ |
| $d = 44$ | $\ell_2$ IP | $0.10 \pm 0.1$ | - |
| (215-269) | $\ell_1$ Rank | $0.19 \pm 0.2$ | $43.90 \pm 0.3$ |
| BCI | $\ell_1$ IP | $0.06 \pm 0.04$ | $\mathbf{71.1 \pm 16}$ |
| $d = 310$ | $\ell_2$ IP | $0.06 \pm 0.04$ | - |
| (288-2592) | $\ell_1$ Rank | $0.07 \pm 0.06$ | $272.0 \pm 7.9$ |

Table 1. Performance of sparse $\ell_1$ infinite push, $\ell_2$ infinite push and sparse SVM Rank on real-world datasets. the column top denotes the average rate of positive examples ranked on top of the list. The column #var depicts the average number of variables used in the models. Underlined performances in the top column highligh methods whose performances are statistically significantly worse than the others according to a Wilcoxon signed rank test at the level of 0.05. Number in parenthesis depicts the number of training and testing examples.

## 5. Conclusions

We have shown in this paper that embedded feature selection based on sparsity-inducing norms can be extended to loss functions that are themselves non-differentiable and intrinsically complex. For sparse SVM infinite push, we have proposed an algorithm based on alternate direction method of multipliers that alternatively solves a Lasso (or related) problem and applies the proximal operator of the infinite push loss. For computing this proximal operator, we have devised a novel algorithm based on the projection on $\ell_1 - \ell_\infty$ ball. Our experimental results show that our sparse

SVM infinite push compares favorably to other approaches in terms of number of variables used in the model as well as in term of accuracy of ranking on top of the list. Future works will focus on algorithms that scale linearly or sublinearly with $m \cdot n$ and on theoretical analysis of the methods.

# Acknowledgments

This work is partially supported by the PASCAL2 Network of Excellence, ICT-216886, ANR Project ASAP ANR-09-EMER-001.